\documentclass[11pt,a4paper]{article}
\usepackage{times,latexsym}
\usepackage{url}
\usepackage[T1]{fontenc}
\usepackage[utf8]{inputenc}
\usepackage[normalem]{ulem}
\usepackage[acceptedWithA]{tacl2018v2}

\usepackage{ocr}
\usepackage{microtype}

\usepackage[smallerops,vvarbb]{newtxmath}

\usepackage{xspace}

\usepackage{xfrac}

\usepackage{algorithm}
\usepackage[noend]{algpseudocode}

\algnewcommand{\LineComment}[1]{\State \(\triangleright\) #1}

\algblockdefx[CommentHead]{CommentHead}{EndCommentHead}[1]{\textcolor{gray}{\(\triangleright\) #1}}{}
\algtext*{EndCommentHead}

\algblockdefx[Class]{Class}{EndClass}[1]{\textbf{class} #1}{}
\algtext*{EndClass}

\algnewcommand\algorithmicbreak{\textbf{break}}
\algnewcommand\Break{\State \algorithmicbreak}
\algnewcommand\algorithmiccontinue{\textbf{continue}}
\algnewcommand\Continue{\State \algorithmiccontinue}

\algnewcommand\algorithmicinput{\textbf{Input:}}
\algnewcommand\algorithmicoutput{\textbf{Output:}}
\algnewcommand\Input{\item[\algorithmicinput]}
\algnewcommand\Output{\item[\algorithmicoutput]}

\DeclareMathAlphabet{\mathcal}{OMS}{cmsy}{m}{n}

\usepackage{url}
\usepackage{colortbl}

\usepackage{tikz}
\usetikzlibrary{arrows,matrix,positioning,fit,calc,shapes,decorations.pathreplacing,svg.path,patterns}
\pgfdeclarelayer{background}
\pgfdeclarelayer{backgroundoverlay}
\pgfsetlayers{background,backgroundoverlay,main}
\usepackage{pgfplots}
\pgfplotsset{compat=1.13}

\usepackage{nicefrac}
\usepackage{enumerate}

\usepackage{adjustbox}

\newcommand{\charword}[1]{\scalebox{.9}{\ocrfamily #1}\xspace}

\usepackage[capitalise]{cleveref}

\crefname{section}{\S}{\S\S}
\Crefname{section}{\S}{\S\S}
\crefformat{section}{\S#2#1#3}  

\title{Between words and characters:\\ A Brief History of Open-Vocabulary Modeling and Tokenization in NLP}

\author{
    Sabrina J. Mielke $^{1,2}$ \qquad Zaid Alyafeai $^3$ \qquad Elizabeth Salesky $^1$ \\
    \bf Colin Raffel $^2$ \qquad Manan Dey $^4$ \qquad Matthias Gallé $^5$ \qquad Arun Raja $^6$ \\
    \bf Chenglei Si $^7$ \qquad Wilson Y. Lee $^8$ \qquad Benoît Sagot $^9$\thanks{\; Working group chairs} \qquad Samson Tan $^{10}$\footnotemark[1] \\
    \textit{BigScience Workshop Tokenization Working Group} \\[.5em]
    \small
    $^1$Johns Hopkins University \;\;
    $^2$HuggingFace \;\;
    $^3$King Fahd University of Petroleum and Minerals \;\;
    $^4$SAP \\
    \small
    $^5$Naver Labs Europe \;\;
    $^6$Institute for Infocomm Research, A*STAR Singapore \;\;
    $^7$University of Maryland \\
    \small
    $^8$BigScience Workshop \;\;
    $^9$Inria Paris \;\;
    $^{10}$Salesforce Research Asia \& National University of Singapore \\
    \texttt{sjm@sjmielke.com}
}

\date{}

\begin{document}
\maketitle

\pagestyle{plain}
\thispagestyle{plain}

\begin{abstract}
What are the units of text that we want to model?
From bytes to multi-word expressions, text can be analyzed and generated at many granularities.
Until recently, most natural language processing (NLP) models operated over words, treating those as discrete and atomic tokens, but starting with byte-pair encoding (BPE), subword-based approaches have become dominant in many areas, enabling small vocabularies while still allowing for fast inference. Is the end of the road character-level model or byte-level processing?
In this survey, we connect several lines of work from the pre-neural and neural era, by showing how hybrid approaches of words and characters as well as subword-based approaches based on learned segmentation have been proposed and evaluated.
We conclude that there is and likely will never be a silver bullet singular solution for all applications and that thinking seriously about tokenization remains important for many applications.
\end{abstract}

\section{Introduction}

\begin{quote}
    \emph{```tokens' are not a real thing. they are a computer generated illusion created by a clever engineer}'' \hfill---\texttt{@dril\_gpt}\footnote{A Twitter bot with (human-curated) outputs of a language model based on GPT2 \citep{radford2019language} and trained on tweets of Twitter poet \texttt{@dril};  \url{https://twitter.com/dril_gpt2/status/1373596260612067333}.}
\end{quote}

When we first introduce people to NLP models, we often take for granted the idea that text is cut up into little pieces that are fed to a computer, eventually as nothing but a sequence of integers.
Following \citet{webster-kit-1992-tokenization}, we call these (usually) contiguous substrings \emph{tokens}. In teaching settings and in fact historically in NLP, these tokens are somewhat naturally implied to be \emph{words}---at first, perhaps naively as ``space-separated substrings'' in English. Understandably, as soon as we try to split punctuation from words, things get a little tricky.
Take the word \emph{``don't''} for example: not splitting it is reasonable, but if we split on all punctuation, we would get three somewhat nonsensical tokens (\emph{don ' t})---and we might argue that the most sensible split actually ought to yield the two units ``\emph{do}'' and ``\emph{n't}'', as found in the Penn Treebank \cite{marcus-etal-1993-building}.

\begin{figure}[t]
    \centering
    \includegraphics[width=\linewidth]{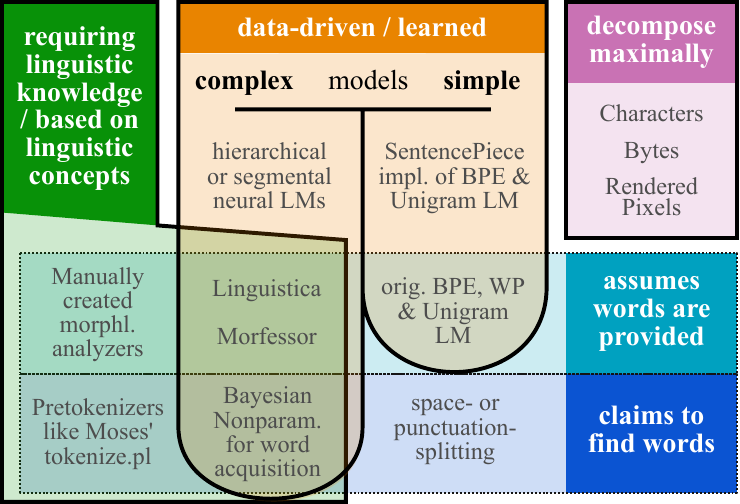}
    \caption{A taxonomy of segmentation and tokenization algorithms and research directions}
    \label{fig:taxonomy}
\end{figure}

This survey deals with such questions of \emph{tokenization} and we will elaborate on fundamental questions and terminology in \cref{sec:basics}, and show how this important if somewhat unglamorous part of all NLP work has historically been treated (\cref{sec:pretokenization}). However, since especially in the last five years there has been renewed interest in going \emph{beyond} intuitive definitions of a ``token'' as a somewhat atomic word-like space-separated unit. One way to do so is to use word-internal information to augment word-like units (\cref{sec:word-level}), which neural modeling has made easier than ever, leading to models that can even \emph{learn} word boundaries with no overt indication (e.g., when spaces are present).
That notion of unsupervised word segmentation or discovery has been present on its own in decades of work (\cref{sec:find-words}), but we will find it useful to consider when finally looking at the now prevalent idea of using subword units as atomic tokens (\cref{sec:find-subwords}).
We will close out this survey with a look at some issues of sharing and competition in multilingual vocabularies (\cref{sec:multilingual}) and work using the simplest tokenization possible: maximal decomposition into characters, bytes, or even pixels (\cref{sec:purechar}).

Equipped with all this knowledge, we will conclude the survey in \cref{sec:conclusion} by making a case for why ``complicated'' tokenization is something to practice or even learn about even now in 2021 when the easy solution of bytes seems in reach. We will argue that while recent advances like CANINE \citep{clark2021canine}, ByT5 \citep{xue2021byt5}, or Charformer \citep{tay2021charformer} make maximally decomposed processing feasible for certain domains and use-cases, they do not cover the wide variety of NLP scenarios and come with their own drawbacks and biases.

\section{Tokens, word-forms, and sub-words}\label{sec:basics}

In NLP, textual data has been traditionally segmented into ``sentences'' (or ``utterances'', etc.) and ``words'' due to linguistic motivations and technical constraints. The macroscopic units (``sentences'') are often considered independently from one another and themselves segmented into microscopic units. The definition of these microscopic units has always been a matter of approximation and compromise. On the one hand, these units receive linguistic annotations (e.g.~part-of-speech tags, morphosyntactic annotation, syntactic dependency information), which would require them to be linguistically motivated units. On the other hand, a large range of phenomena make it highly non-trivial to identify and even to consistently define linguistic units, denoted by the Morphological Annotation Framework (MAF) ISO standard \cite{MAF} as {\em word-forms}. Such phenomena include contractions (e.g.~English {\em don't}, cited above, and French {\em aux} `to the\textsubscript{pl}'), compounds (e.g.~French {\em copier-coller} `copy-paste'\footnote{Cf.~inflected forms {\em copié-collé} `copy-pasted', lit.~`copied-pasted', but {\em copie-collerai} `will\textsubscript{1sg} copy-paste'.}), morphological derivatives (e.g.~English or French {\em anti-Trump}), as well as numerous classes of named entities and other sequences following type-specific grammars (e.g.~numbers, URLs).

As a result, {\em typographic units}, generally called {\em tokens} have been used as an approximation for such linguistically motivated units. For instance, MAF defines a token as a ``non-empty contiguous sequence of graphemes or phonemes in a document.'' In the case of writing systems using a typographic separator such as the whitespace, now universally used with the Latin script for instance, {\em tokens} have been widely used and broadly defined as either contiguous sequences of non-punctuation non-whitespace marks or punctuation marks. Provided a few arbitrary decisions are made regarding certain punctuation marks (e.g.~the hyphen or the apostrophe), such a definition makes it possible to deterministically split a sentence into atomic units, resulting in a segmentation into tokens that are acceptable approximations of word-forms. Crucially, as discussed in detail by \citet{MAF}, \citet{sagot:inria-00515489} and elsewhere, there is no one-to-one correspondence between tokens and word-forms; a word-form can be made of several tokens (e.g.~French or English {\em sine die}) whereas several word-forms can be represented by the same token (e.g.~English {\em don't} = {\em do + not}, Spanish {\em damélo} = {\em da + me + lo}). This is what the Universal Dependencies guidelines\footnote{\url{https://universaldependencies.org/u/overview/tokenization.html}} refer to as ``multitoken words'' and ``multiword tokens,'' respectively, a topic further discussed by \citet{more-etal-2018-conll}. In fact, both phenomena can interfere in non trivial ways (e.g.~French {\em à l'instar du} = {\em à\textunderscore l'instar\textunderscore de + le}).\footnote{When the writing system at hand does not have a typographic separator, tokens must be defined differently. With scripts like the Chinese or Japanese script, an option for instance is to consider each character as a token on its own.}

In recent years, the spread of approaches based on neural language models resulted in an evolution in how sentences are split into atomic units, thereby resulting in a redefinition of the notion of tokenization. Indeed, based both on scientific results (e.g.~the impact of sub-word segmentation on machine translation performance \citep{SenHadBir16Neural}) and on technical requirements (e.g.~language models such as BERT that require a fixed-size vocabulary), the need for the atomic processing units (still called tokens) to be an approximation of word-forms has faded out. As a result, in current NLP, the notion of token still perfectly matches its MAF definition, but it no longer corresponds to the traditional definition of a typographic unit. ``Tokenization'' now denotes the task of segmenting a sentence into such non-typographically (and indeed non-linguistically) motivated units, which are often smaller than classical tokens and word-forms, and therefore often called {\em sub-words}. Typographic units (the ``old'' tokens) are now often called ``pre-tokens,'' and what used to be called ``tokenization'' is therefore called nowadays ``pre-tokenization.'' This term is motivated by the fact that the first approaches to the new notion of ``tokenization'' often involved segmenting sentences into proper typographic units (i.e.~the ``old'' notion of tokenization) before further segmenting (some of) the resulting units (formerly ``tokens'', now ``pre-tokens'') into ``sub-words''.

\section{Pre-tokenization yields word-like typographic units}\label{sec:pretokenization}

As a compromise between the linguistic irrelevance of purely typographic tokens and the difficulty of automatically splitting a text into linguistically motivated word-forms, units that are halfway between purely typographic tokens and purely linguistic word-forms have been widely used,\footnote{Cf.~the Penn TreeBank \cite{marcus-etal-1993-building}, where {\em don't} is split into two units {\em do} and {\em n't}, as mentioned above.} albeit often (improperly) denoted by the term ``token'' before the spread of sub-word tokenization, and ``pre-token'' since then. Many tools, formerly known as ``tokenizers'' and nowadays as ``pre-tokenizers'', have been developed and used for a long time. Some of them are relatively simple and remain faithful to the typographic token. Amongst the most widely used, we can cite the venerable Moses \citep{koehn-etal-2007-moses} tokenizer%
\footnote{\url{https://github.com/moses-smt/mosesdecoder/blob/master/scripts/tokenizer/tokenizer.perl}}
and the more recent \texttt{pre-tokenizers} package in Hugging Face's Tokenizers package.%
\footnote{\url{https://github.com/huggingface/tokenizers}} 
This practice resulted in the word ``tokenization,'' now ``pre-tokenization,'' ending up denoting the task of segmenting sentences into atomic units in general, i.e ``basic units which need not be decomposed in a subsequent processing'' \cite{webster-kit-1992-tokenization}, even when such units are closer to word-forms than to typographic units.

Moreover, it is fairly common for tokenizers to not only segment sentences but also modify the raw text, for instance for \emph{normalization}, spelling correction or named entity detection purposes, thereby departing from the standard definition of token. Thus a string like ``some `quoted' text'' might be tokenized into five units: ``some " quoted " text.'' On the other hand, a number of tools have been developed and used that attempt a joint segmentation of sentences into tokens (which remain proper substrings of the input text) and word-forms (which can be the output of normalization, spelling correction and named entity recognition steps).
Some of these tools take into account the inherent non-determinism of the mapping between the two types of units \cite{sagot:inria-00515489}.
Normalization operations like these or conflation and merging of different whitespace symbols leads to most tokenizers being \emph{irreversible}, i.e., we cannot recover the raw text definitively from the tokenized output.\footnote{While this is not a major issue for most applications, it means that we no longer model the original text, but a string that may correspond to many original strings, inflating probability estimates of language models; this issue is also highlighted in the context of ambiguous tokenization (see \cref{sec:unigramlm}) by \citet{cao-rimell-2021-evaluate}.}$^,$%
\footnote{The ``reversible'' language-agnostic tokenizer of \citet{MieEis18Spell} attempts to remedy some of these issues, but still conflates whitespace.}

\section{Augmenting word-level pretokenizer tokens with character information}\label{sec:word-level}

While word-level models are conceptually easy to understand and in the neural era \citep{bengio2001lm,collobert2008unified,mikolov2013distributed} offer features at an interpretable granularity, their central weakness is the inability to deal with rare and novel words, i.e., words that were seen very rarely during training or not even at all (out-of-vocabulary, OOV)---they are \emph{closed-vocabulary} models.
In particular, historically rare word types were replaced with a new word type \textsc{unk} (\emph{unknown}) at training time; at test time, any token that was not part of the model's vocabulary could then be replaced by \textsc{unk}.
That approach however comes with a number of drawbacks: 1) \textsc{unk}s are not acceptable when performing natural language generation (NLG), 2) they do not allow us to extract features for novel words that are useful anchors of meaning and not just one-off events \citep{Chu00Empirical} when used in large-scale models like ELMo \citep{PetNeuIyy18Deep} or BERT \citep{DevChaLee18Bert}, and 3) in languages other than English, in particular those with more productive morphology and thus higher type-token-ratio, removing rare words is infeasible \citep{cotterell-etal-2018-languages,mielke-etal-2019-kind}.
Nevertheless, since the word \emph{is} a fundamental unit of language, a number of approaches emerged to improve handling of rare and novel words under a fundamentally word-based framework by basing their handling on the characters that make up a word.
We will present some of these approaches in this section, for a more comprehensive treatment of word representation, \citet{pinter2021integrating} surveys linguistic background and multiple approaches.

\subsection{Augmenting word-level models with spelling information}\label{sec:augmenting}

The idea of somehow using information about the spellings of a word to inform the word's representations of course is decades old.
In neural models for language, research in the 90s and 2000s often forewent the focus on words altogether and processed strings of characters instead (see \cref{sec:purechar} and \cref{sec:related-twolevel}), but as soon as neural models became important in NLP, combinations of word- and character-level information for use in neural networks emerged there, too.

\Citet{SanZad14Learning} first proposed to use information about the words themselves to aid word embedding estimation.
Soon thereafter \citet{LinDyeBla15Finding}, \citet{KimJerSon16Character}, and \citet{45446} popularized the idea of deterministically constructing a word's embedding from its spelling,%
\footnote{It should be noted that \citet{45446} also propose a variant in which output tokens are not scored through a softmax, but generated character by character, anticipating the advancements described in \cref{sec:related-twolevel}, but still staying in a closed-vocabulary setup.}
both for textual input as well as for generative language modeling, that is, prediction of strings.
 However, even when replacing embedding matrices with convolutional neural network (CNN) layers, their generative models are still \emph{closed-vocabulary}, meaning they can only predict words that were seen (often enough) in training data, so the CNN construction only helps with rare words, not novel words.
Furthermore, constructing embeddings from spellings for each token (as opposed to every type like \citet{MieEis18Spell}, see \cref{sec:related-twolevel}) implicitly trains the CNN-powered embedding function to ``get frequent words right'' instead of anticipating novel words, an issue discussed in \citet{MieEis18Spell}.
Similar constructions led to advances in other classic NLP tasks like POS tagging \citep{plank-etal-2016-multilingual} and ultimately powered the first big \emph{contextual word embedding} model ELMo \citep{PetNeuIyy18Deep}.

The popular fastText embeddings \citep{TACL999} propose constructing word embeddings not from characters, but from overlapping $n$-grams, allowing one to obtain embeddings for novel words (making it ``open-vocabulary'' in that sense, though not in the generative sense).
\citet{ataman-federico-2018-compositional} likewise obtain better performance on machine translation by using (overlapping) $n$-grams instead of characters (also beating BPE on morphologically rich languages).

In more recent times, \citet[][CharacterBERT]{el-boukkouri-etal-2020-characterbert} and \citet[][CharBERT]{Ma2020CharBERTCP} use the same CNN construction as in \citet{KimJerSon16Character} on a modern BERT-style model, this time enhancing the BPE units' embedding with their constituent characters' embedding,
motivated by better handling noisy texts with spelling errors or transferring to new domains like medical text; concurrently, \citet{aguilar2021char2subword} do almost the same, but using a small Transformer instead of CNNs.

Finally, construction-based approaches have also been integrated into pretrained word-level input models. Specifically, \citet{pinter2017mimicking} learn a model that is trained to \textit{mimic} the embedding of a word given its spelling using a helper RNN model that is called whenever an unknown word appears during test time.

\subsection{Open-vocabulary language modeling with (tokenizer-defined) words made of characters}\label{sec:related-twolevel}

Extending closed-vocabulary generative models to open-vocabulary models, i.e., those that can predict and generate novel words at test time, is somewhat more difficult than being open-vocabulary on the \emph{input} side because it must be possible to hold out probability mass for the infinite set of sentences that contain completely novel words.

Inspired by \citet{LuoMan16Achieving}, \citet{MieEis18Spell} propose a probabilistic two-stage model that essentially augments the ordinary closed-vocab word-level recurrent neural network language model (RNNLM) setup by regularizing word embeddings to be predictive of their spellings using a smaller character-level RNNLM and using that smaller model to generate novel words on the fly whenever the word-level RNNLM predicts \textsc{unk}, yielding an open-vocabulary model motivated by linguistic notions and intuitive modeling and proven successful qualitatively and quantitatively.

Independently developed, the model of \citet{KawDyeBlu17Learning} follows a similar two-level setup of word- and character-level RNN, but where each word has to be spelled out using a character-level RNN if it cannot be directly \emph{copied} from the recent past using a cache model \citep{grave2016improving}.%
\footnote{As mentioned before, the idea of spelling out words in isolation from hidden states had previously proven unsuccessful in \citet{45446}'s comparison, but this was in a closed-vocab setup and without the caching mechanism \citet{KawDyeBlu17Learning} employ.}
Their analysis shows clearly that the cache model not only copies ``bursty'' unknown words like \charword{Noriega} \citep{Chu00Empirical}, but also extremely common function words like \charword{the} in an attempt to keep itself from forgetting them.
The idea is picked up by \citet{ataman-etal-2019-importance} for a machine translation decoder (creating word embeddings on the encoder side from character-level BiRNNs as in ELMo \citep[see \cref{sec:augmenting}]{PetNeuIyy18Deep}) and later extended by \citet{Ataman2020A} with some additional stochasticity that is intended to pick up on lemmata and inflections unsupervisedly.

A different approach is having higher layers of multi-layer RNNs run at lower speed (skipping updates to the hidden state)
This is an old idea, first present in \citet{ElHBen95Hierarchical} (building on \citet{schmidhuber1991neural,schmidhuber1992learning}'s ``neural sequence chunker'') and revived in \citet{pmlr-v32-koutnik14} for fixed-frequency skipping and \citet{HwaSun17Character} for skipping on word boundaries (which are assumed to be observed).%
\footnote{Specifically, \citet{HwaSun17Character} describe an architecture in which character-level and word-level models run in parallel from left to right and send vector-valued messages to each other.  The word model sends its hidden state to the character model, which generates the next word, one character at a time, and then sends its hidden state back to update the state of the word model.}
This approach leads to the first of a number of ways in which we can actually \emph{learn} word boundaries and thus segmentations.

\section{Learning segmentations to find concatenative word-like pretokenizer tokens}\label{sec:find-words}

So far we have relied on having a predefined notion of word (or pretokenization output) despite the conceptual struggles outlined in \cref{sec:basics}. But what if such a definition is not given, not obtainable, or simply not desirable (for reasons of robustness and in languages other than English etc.)?
Is there a way to let our data-driven machine learning approach also \emph{learn} the tokenization?
Most approaches described in this section propose to tackle tokenization by treating the implied \emph{segmentation} as a latent variable (with an exponentially-sized domain) on which we can perform approximate or (using more assumptions) exact inference to find segments and boundaries that hopefully correspond to meaningful units. The various techniques described in this section yield units of varying size and quality.

\subsection{Character-level neural models that learn to skip steps at higher levels}\label{sec:segmental}

Already in the 90s, \citet{elman1990} manually analyzed character-level RNNs and correlated their prediction surprisal with word boundaries. This idea that was then expanded on in \citet{schmidhuber1991neural,schmidhuber1992learning}'s ``neural sequence chunker''.
More recently, surprisal was applied to not only character-level neural models but also n-gram models under a beam search framework by \citet{doval-gomezrodriguez} to split microblog texts in which spaces are deleted.

Instead of using post-hoc surprisal thresholding, the HM-RNN \citep{ChuAhnBen17Hierarchical} takes the idea of multiple timescales motivated in \cref{sec:related-twolevel}, but \emph{learns} the binary decision to skip or update (thereby providing a sense of word boundaries), optimizing with approximate gradient descent using the straight-through estimator \citep{DBLP:journals/corr/BengioLC13}. In their model, communication between layers happens bidirectionally: the lower network reports its final state to the higher one; that higher network reports its new state to the lower layer that then proceeds to run by itself and so on.
While they ``recover'' word boundaries when including spaces in their data, \citet{kawakami-etal-2019-learning} claim to get unusable segments with this model when not including spaces. Furthermore, when trying to use the HM-RNN for NMT, \citet{CheFosBap18Revisiting} report that it took a lot of fixing to get it to train at all; its performance on the task was competitive but not superior.
This finding corroborates that of \citet{KadCotChr18Revisiting}, who dedicate a paper to trying to get the HM-RNN to train well, ablating it, and also showing subpar segmentations on text data (as well as the worrying inability to reach the original reported numbers).
\citet{KreSok18Learning} try to use a similar paradigm of skipping steps and generating summaries with lower layers for NMT and find (similarly to \citet{KadCotChr18Revisiting}) that skipping is rarely used and thus seems to be unnecessary for good performance.
Nevertheless, the model is extended to phrase- and sentence-level boundaries by \citet{DBLP:journals/corr/abs-2106-02562}.

It is worth pointing out that despite having coarser layers of computation, these models still have to ``spell out'' a word every time it is generated, i.e., they cannot \emph{memoize} tokens as reusable units.

\subsection{Marginalization over all possible segmentations}

Finally, a conceptually straightforward approach is to treat the segmentation of a string as a latent variable that needs to be marginalized over both at training and test time. This essentially means having a vocabulary that contains strings of differing lengths that overlap, i.e., it may contain ``cat,'' ``at,'' and ``foster cat,'' such that the string ``my foster cat'' can be decomposed a number of ways corresponding to different sequences of latent units.
As the number of segmentations is exponential in the sequence or context length, we need to either resort to approximations for marginalizing over latent decompositions (\cref{sec:approximate_segmentation_marginalization}) or simplify the model with independence assumptions e.g. by using an $n$-gram model (\cref{sec:snlm}).

\subsubsection{Approximate marginalization}
\label{sec:approximate_segmentation_marginalization}

\citet{ChaZhaLe17Latent} propose an estimator to approximate the marginal probability of observations using approximate MAP inference through beam search.
They find that the model is very hard to train, but manage to obtain promising results.
\citet{BucNeu18Neural} confirm this model's instability and propose some approximate inference schemes based on averaging RNN hidden states that produce better results in terms of LM perplexity.
\citet{hiraoka-etal-2020-optimizing} implement a similar model, based on a Unigram LM tokenization proposal distribution (see \cref{sec:unigramlm}), whose $n$-best tokenizations of a sentence are fed into any sentence encoder model independently and whose resulting sentence embeddings are averaged in line with their a priori tokenization likelihood. \citet{hiraoka-etal-2021-joint} extend this model to sequence-to-sequence settings by training a tokenizer and downstream model with separate losses, the former by rewarding tokenizations that produced a low downstream loss, and the latter using just one tokenization sampled from the conditioned (and tempered) LM.

\subsubsection{Exact marginalization using additional independence assumptions: segmental neural language models}\label{sec:snlm}

The more popular solution of \emph{segmental neural language models} was pioneered by \citet{KonDyeSmi16Segmental}, who cast the problem of segmentation as a monotonic%
\footnote{Interestingly, with some reordering of the input one can break monotonicity between input and output, making the model similar to phrase-based MT \citep{HuaWanHua18Towards}.}
seq2seq task, going from characters to a covering sequence of substrings, i.e., a segmentation. By conditioning segment prediction on the \emph{entire raw string}, processed and embedded using a BiRNN, segmentation decisions/scores can use context, but by scoring every individual possible substring independently as a segment using these embeddings and then adding up individual scores to score entire segmentations, they can find a covering of the entire input string with segments efficiently using dynamic programming.
The reason for this ability is the central independence assumption: the model does not depend on any \emph{other segments} when scoring a segment, but merely on surrounding \emph{characters}.
\citet{WanWanHua17Sequence} extend this by also having a per-segment RNN over characters for the outputs that can run without knowing the segmentation and whose past representations can thus be used by the individual segment generation processes, allowing for left-to-right sharing of information about segments without breaking dynamic programming.

The jump to LMing is now made simply by omitting the conditioning on an input, yielding the model of \citet{sun-deng-2018-unsupervised}, who coin the term segmental language model, training on Chinese characters and using the unsupervisedly learned segments to compete on Chinese Word Segmentation. To keep the computation of the quadratic number of segments feasible, they restrict the segments to a maximum length of 4 characters (a sensible prior for Chinese).
\citet{grave-etal-2019-training} make the same jump independently, using Transformers as the independent character-level global backbone. When evaluating on English open-vocabulary language modeling, \citet{grave-etal-2019-training} notice improved perplexity, but not using or evaluating the obtained segmentation, most likely because they, too, only use 4-grams that appear at least 400 times.
Contemporaneously, \citet{kawakami-etal-2019-learning} use the same independence idea, but have emissions of string segments come from a context-dependent \emph{mixture} of a character-level model like in \citet{KawDyeBlu17Learning} (see \cref{sec:related-twolevel}) and a large set of substrings (up to 10-grams that appear often enough in training data) with learned embeddings. They evaluate not only on perplexity, but also on word segmentation performance, where they do beat some baselines (see \cref{sec:nonparametrics}), but still perform much worse than some previous models,%
\footnote{On English they cite the pre-neural models of \citet{johnson-goldwater-2009-improving} and \citet{berg-kirkpatrick-etal-2010-painless} as significantly better; on Chinese, they are beaten by pre-neural models like the one of \citet{MocYamUed09Bayesian} and the neural model of \citet{sun-deng-2018-unsupervised}. More information about some pre-neural models is given in \cref{sec:nonparametrics}.}
which they argue tuned their hyperparameters on segmentation performance instead of marginal likelihood and thus have an unfair advantage.

Interestingly, when training on image captions, \citet{kawakami-etal-2019-learning} find that both perplexity and segmentation performance improve when the model also has access to the image that is being described, showing that learning segmentation only from monolingual and unimodal text may be harder than when other modalities or languages are present.
This observation is shared by \citet{he-etal-2020-dynamic}, who build a similar segmental model (in their case, a Transformer-based version that still follows the character backbone idea to allow for dynamic programming) as the target-side generator in an NMT system and use it not as the final model, but merely as a learned tokenizer. This is easy to achieve by changing the dynamic program from marginalization to maximization and thus obtaining a new segmentation, called DPE, that can be used in place of BPE or unigram LM (see \cref{sec:bpe-and-friends}).
\citet{he-etal-2020-dynamic} proceed to show that learning to tokenize with a small Transformer-based NMT model%
\footnote{Unlike previously mentioned papers, they however restrict the vocabulary to units of an input BPE vocabulary instead of using length and frequency heuristics.}
produces better segmentations than BPE for use in a bigger model; in particular, training the tokenizing model on the translation task produces different segmentations depending on the source language, and, more importantly, better segmentations (as measured through downstream translation performance) than training on target-side language modeling alone.

The idea of conditioning on characters and predicting segments is extended to the adirectional masked language modeling setting found in Transformers and left-to-right autoregressive Transformers by \citet{downey2021masked}, though results do not outperform RNN-based SNLMs consistently.

Note that many of these models can also be seen as relatives of models based on UnigramLM, which we will cover in \cref{sec:unigramlm}.

\subsection{Finding words through Bayesian non-parametrics}\label{sec:nonparametrics}

In the era of $n$-gram and word-based language models, \citet{MacPet95Hierarchical} noticed that a Bayesian view of autoregressive language models may prove beneficial, reinterpreting smoothing and backoff in $n$-gram models as inference in a hierarchical model where higher-order distributions are drawn from a Dirichlet distribution whose mean is a lower-order distributions.
\Citet{Teh06Hierarchical} extends this thinking, proposing a \emph{hierarchical} PYP language model where we again have $n$-gram distributions of arbitarily large orders, drawn through a hierarchy of PYP distributions that lead to a model that still bears resemblance to $n$-gram language model smoothing, but offers a principled way to forego the choice of $n$. The pinnacle of this idea of modeling was reached in \citet{WooGasArc11Sequence}'s sequence memoizer, which boasted great compression performance for arbitrary binary data and still performed very well on language modeling tasks, although neural models at this time already proved to be strong competitors.

At the same time, \citet{goldwater2006powerlawgenerators} extended this Bayesian perspective to also explain how new words are first coined and how they are then used in running text: a process they call two-stage language modeling (see \cref{sec:related-twolevel}), with the two stages being referred to as \emph{generator} (which creates new lexemes) and \emph{adaptor} (which governs reuse; here, a Pitman-Yor Process (PYP)), relating the resulting interaction between types and tokens to interpolated Kneser-Ney smoothing as presented in \citet{chen-goodman}.%
\footnote{The formalism of generators and adaptors is extended and formally specified under the name \emph{adaptor grammars} in \citet{johnson-et-al-2007} and used very successfully for state-of-the-art word segmentation in \citet{johnson-goldwater-2009-improving}.}
Given such a two-stage model to explain text and the use of Bayesian nonparametrics that can assign positive probability to an infinite number of possible lexemes, it becomes possible to also try to infer word boundaries, that is to perform unsupervised word segmentation. Motivated more by trying to explain and model cognitive processes and in particular child language acquisition, \citet{goldwater2009bayesian}%
\footnote{The idea and partial results are already presented in \citet{goldwater-et-al-2006}, but the authors request citing the updated 2009 paper. \citet{GolGriJoh11Producing} summarized this thread of research.}
summarize Unigram and Bigram Dirichlet Processes (DPs) for segmenting transcribed infant-directed speech, showing superiority over older non-Bayesian approaches.
\citet{MocYamUed09Bayesian} extend the idea from bigram DPs to $\infty$-gram nested/hierarchical PYPs to improve word segmentation for English written text; \citet{elsner-etal-2013-joint} additionally model phonotactic processes that convert a sequence of segments into observed realizations.

\subsection{Related task: Unsupervised Chinese Word Segmentation}\label{sec:chinesewordsegmentation}

Word segmentation for languages without white-space delimiters such as Chinese, Japanese and Vietnamese \citep{shao2018universal} is an important area of research and can be notoriously tricky.

In Chinese word segmentation (CWS), there is growing interest in exploring unsupervised word segmentation involving neural language models.
Traditionally, popular unsupervised approaches take on two primary paths: 1) discriminative models and 2) generative models. Discriminative models rely on goodness measures for candidate segmentation. These statistical measures incude Mutual Information (MI), normalized Variation of Branching Entropy (nVBE) and Minimum Description Length (MDL), etc., see \cref{sec:morfessor}.
Generative models focus on designing statistical models to find the optimal segmentation of the highest generative probability. These models include Hidden Markov Model (HMM), Hierarchical Dirichlet Process (HDP), Nested Pitman-Yor Process (NPY), etc., see \cref{sec:nonparametrics}.
It is trivial to extend discriminative approaches by replacing $n$-gram language model with neural language models. For generative approaches, previous work has shown that constructing a neural language model with a context encoder and a segment decoder achieves competitive performance to its statistical counterparts \citep[][see previous subsection \cref{sec:snlm}]{sun-deng-2018-unsupervised}.

\section{Learning subword vocabularies and segmentations}\label{sec:find-subwords}

As teased in \cref{sec:pretokenization}, \emph{subword} units allow for a smooth transition between a word-level model and a character-level model: split the word-like tokens obtained by pre-tokenization into smaller units: the set of all possible subword units is finite and can be determined from training data, but it is assumed to include all characters (or bytes, see \cref{sec:bytes}) that appear at test time, making it possible to explain any novel word in held-out data.

While thinking about subword information may have more tradition for processing morphologically rich languages, \Citet{MikSutDeo12Subword} already proposed using \emph{subword units}\footnote{They don't supply many details, but claim that the units they use are \emph{syllables}---and that they help.} instead of words for language modeling English to avoid out-of-vocabulary (\emph{OOV}) issues.
Since \citet{SenHadBir16Neural}, however, it has become customary in many if not most current-day NLP models to combat large and infinite vocabulary size.

What then should we use as subword units? One option are manually constructed, linguistically informed rule-based systems (\cref{sec:fsts}), another is given by data-driven segmentation learners, which traditionally have been motivated and evaluated either linguistically (\cref{sec:morfessor}) or given by simple heuristics to be fast and easy and improve downstream performance (\cref{sec:bpe-and-friends}).

It is important to point out that despite reasonable motivation, segmentation may be a bad idea in for example Semitic languages like Arabic and Hebrew \citep{DBLP:journals/corr/abs-1809-01301} or other languages with non-concatenative morphological phenomena, which \citet{amrhein-sennrich-2021-suitable-subword} claim are better served by character-level models or those with very small subword inventories.

\subsection{Manually constructed linguistic analyzers}\label{sec:fsts}

Morphological analysis is of great importance for morphologically rich languages and various tools have been developed to analyze word forms into their lemmata and inflections, earliest and most famous of them the Porter stemmer \citep{Porter1980} for English.
These tools are often constructed manually by linguists using finite-state tools \citep[FST;][]{beesley2003finite} as often morphological processes lend themselves to systematic description, making it faster and cheaper to manually construct finite-state analyzers than to try to learn complicated data-driven models \citep{beemer-etal-2020-linguist}.

Interestingly, such finite-state tools can not only be used for overt segmentation or discovery of lemmata and other subword units, but even a morphological tagger can be used to induce segmentations, as finite-state machines allow for easy tracking of what parts of an input string resulted in output choices, allowing one to identify for example affixes.

It is worth pointing out that such analysis, yes, relies on manual annotation, but beyond that is often considered slow and needlessly complicated. Nevertheless combinations of lemmatization of tagging have been used successfully to tackle large and potentially noisy vocabularies, for example by \citet{tan-etal-2020-mind}'s BITE, which converts inflected forms into lemma and tag to protect against noise and improve on dialectal data.

An important difference to the rest of this survey is that such an approach has the potential to be stronger even, as foregoing purely concatenative segmentation allows one to ``segment'' for example the word ``hoping'' as ``hope V.PTCP;PRS'' or ``ate'' as ``eat PST,'' allowing sharing of information with other forms in the respective paradigm. The benefit of such an approach is also shown by \citet{hofmann-etal-2021-superbizarre}, who observe that undoing derivational processes by splitting words into morphemes before tokenizing can improve sentiment and topicality classification results.

\subsection{Other Language-Specific Methods}

German, where compounds are never separated with spaces, has prompted research into compound splitting \citep{koehn-knight-2003-empirical,Braschler2004,macherey-etal-2011-language}.
Another tricky example is Sanskrit, where segmentation is complicated by the fact that there are processes that occur at / cross word boundaries \citep{krishna-etal-2017-dataset,Huet03lexicon-directedsegmentation}.
More recently, in the era of neural and subword-based models, questions of tokenization have most recently been researched for Arabic, where \citet{Alyafeai2021EvaluatingVT} examined various language-agnostic and language-specific tokenizers and find that the performance varies depending on factors like the size and morphological complexity of the datasets.
For Chinese, \citet{Si2021SHUOWENJIEZILI} converted characters into stroke orders or romanization sequences before applying BPE in order to capture potential sub-character information based on glyph or pronunciation. 
\citet{Park2020AnES} shows that a hybrid approach of morphological segmentation followed by BPE (\cref{sec:bpe-and-friends}) works best for most Korean datasets.

\subsection{Unsupervised morphological segmentation}
\label{sec:morfessor}

While subword representations are now commonly evaluated based on their use for a downstream application, initial work in this space often directly assessed the \textit{linguistic} validity of subword\footnote{The resulting units are often termed `morphs' in such settings, representing the surface forms of morphemes.} segmentations by means of databases such as CELEX \citep{celex1995} or annotations from linguistic grammars and analyzers.

In both computational models of corpora and speakers, it has been found that ``distributional regularity and phonotactic constraints are useful for segmentation'' \citep{brent1996distributional}. 
\citet{demarcken-1996-acl} proposed to deconstruct text recursively from sentences over words and morphs into characters through ``composition and perturbation,'' presenting results towards recovering linguistically plausible words. 
\citet{brent1995discovering} proposed an essentially minimum description length \citep[MDL;][]{mdl} based approach to morphological segmentation, in which the sum of potential vocabulary units' length and the length of text encoded with this vocabulary is minimized. 
MDL-based approaches were prolifically adapted and expanded for unsupervised morphological segmentation, as in \emph{Linguistica} \cite{goldsmith2001unsupervised}, and found to generate segmentations with high correlations to morpheme boundaries on English and Romance languages \cite{baroni2000distributional,goldsmith2001unsupervised}.
Initially, these approaches were only lightly guided by additional information about possible morphological structure or paradigms---partitioning word types into sets of stems with either suffixes\footnote{Termed \textit{`signatures'} by \citet{goldsmith2001unsupervised}} or prefixes, they could not recursively split morphs into additional subwords or account for conditioned character changes---and so with only one morpheme boundary they were most appropriate only for the languages on which they were initially tested. 

The use of morphological `categories' and additional structure within segmentation models expanded their recall and applicability. 
The \emph{Morfessor} family%
\footnote{A Python implementation of the Morfessor algorithms is provided by \citet{smit-etal-2014-morfessor}.}
comprises several unsupervised and semi-supervised segmentation models which aimed to incorporate linguistic biases to improve initial na\"ive MDL models. 
\emph{Morfessor 1.0} \cite{creutz2002unsupervised}, later called the \emph{Morfessor Baseline}, is a recursive MDL model based on unigram morph frequencies and lengths; without additional structure it has a clear tendency to over-segment and create spurious splits such as `s + plit.'  
\emph{Morfessor CatMAP} \citep{creutz2005inducing}, or categories maximum-a-posteriori, added a hierarchical HMM-based model of the sequential nature of loose morphological categories (prefixes, stems, and suffixes), where priors could be learned in a semi-supervised way from wordlists; this model remained ideal for the concatenative morphology found in such languages as evaluated in the Morpho Challenge\footnote{An evaluation campaign from 2005-10 which focused on unsupervised morphological segmentation; see \citet{virpioja2011tal} for evaluation methodology.}---English, Finnish, Turkish, German \cite{kurimo2010sigmorphon}. 
The \emph{FlatCat} model \citep{gronroos2014morfessor} \textit{flattened} this structure, which reduced accuracy under unsupervised conditions but simplified and improved semi-supervised learning. 
The most recent \emph{Morfessor} model, \emph{EM+Prune} \citep{gronroos-etal-2020-morfessor} merges the above tradition with recent techniques, leading to a model that is a strict generalization of Unigram LM \citep[][see \cref{sec:unigramlm}]{kudo-2018-subword}.

Many of the approaches to morphological segmentation implicitly treated the task as \emph{paradigm learning} by incorporating the notion of morphological paradigms and inflection classes. 
Within this perspective, one research arch focused on expanding the limited paradigmatic structure in early MDL models either through explicit rules, clustering, or `chains' \cite{snover2002probabilistic,creutz2005inducing,monson2007paramor,monson2009probabilistic,lignos2010morsel,narasimhan-etal-2015-unsupervised}.
Another focused on improving segmentation by discovering forms with shared paradigms, by inducing morphological relatedness across surface forms and further allowing for e.g., spelling changes to improve discovery of shared structure across irregular forms \cite{schone-jurafsky-2001-knowledge,snover-brent-2001-bayesian,yarowsky-wicentowski-2000-minimally,bergmanis-goldwater-2017-segmentation}. 
While segmentation from this perspective can result in more compact corpus representations, with higher segmentation recall and greater corpus-level consistency, their precision is often lower than with e.g., \emph{Morfessor} or frequency-driven techniques. 
As briefly discussed in \cref{comparing-morphological-segmentation-to-bpe}, use of morphological segmentation as tokenization for downstream tasks provides only inconsistent improvements compared to the lighter-weight techniques of \cref{sec:bpe-and-friends}, with recent work predominantly electing for the simplicity of these approaches.

\subsection{Modern fast subword segmentation algorithms}\label{sec:bpe-and-friends}

As explained earlier, the breakthrough for subword tokenization nowadays considered central was the use of Byte-Pair-Encoding \citep[BPE;][]{gage-bpe} by \citet{SenHadBir16Neural} for machine translation.%
\footnote{In language modeling, \citet{MerSanLar17Multiscale} were the first to apply BPE to language modeling and \citet{MieEis18Spell} show that a BPE-based baseline beat all state-of-the-art and even their proposed model on some languages, but the idea didn't really take off until really put to the test by state-of-the-art models like the GPT models \citep{RadNarSal18Improving} and BERT \citep{DevChaLee18Bert}.}

\subsubsection{BPE \citep{gage-bpe,SenHadBir16Neural}}

BPE is a compression algorithm from a family termed ``macro-schemas'' \citep{Storer1982} in which substrings are replaced with references to them.
The name was coined in \citet{gage-bpe}, although equivalent algorithms have been applied for pattern discovery in natural language~\citep{Wolff1975} and complexity of genetic sequences~\citep{JimenezMontano1984} earlier.%
\footnote{See \citet{galle2019investigating} for more historical connection and corresponding analyses, e.g., its linear-time implementation by \citet{Repair}.}
When learning a tokenization, BPE replaces pairs of adjacent symbols with a new symbol representing that pair, iteratively merging all occurrences of the pair that occurs most often at any given time. At test time, the same procedure of merging can be performed by executing all recorded merges in the order in which they were conducted during training of the tokenization model.

Byte-Level BPE \citep{wang2019neural} applies BPE not to characters, but raw bytes (see \cref{sec:bytes}); it is used in GPT-2 \citep{radford2019language} and other models.
BPE-dropout \citep{provilkov-etal-2020-bpe} is an extension allowing for subword regularization (see \cref{sec:unigramlm}.

\subsubsection{WordPiece \citep{wordpiece}}

A very similar technique had been proposed under the name ``WordPiece'' by \citet{wordpiece} for Japanese and Korean text (where reliance on space-separated tokens is impossible as text is written without spaces), though it is also used in BERT \citep{DevChaLee18Bert} and other models. Unlike BPE, WordPiece doesn't merge the most often co-occuring pair but pairs that increase the likelihood that an $n$-gram based language model trained with this updated vocabulary reaches on data (the fact that only some counts need to be updated in such a model and the use of frequency-based heuristics for selecting and batching of multiple merge candidates keep the process computationally feasible).
To segment text, WordPiece follows a per-word left-to-right longest-match-first strategy, allowing for very fast linear-time processing \citep{song-etal-2021-fast}.

\subsubsection{UnigramLM \citep{kudo-2018-subword}}\label{sec:unigramlm}

\citet{kudo-2018-subword} picks up on the idea of judging subword candidates by evaluating their use in a language model, but it uses a simple unigram language model (hence calling the algorithm \emph{unigram LM}) and iteratively \emph{removes} subword units from a starting vocabulary that contains far more subword units than are desired: on every iteration, the unigram LM is trained using EM and then the lowest-probability items are pruned from the vocabulary---the process is repeated a few times until a desired vocabulary size has been reached.

Interestingly, this probabilistic setup also cleanly models the fact that there are many possible segmentations that are consistent with a given string (in the extreme case, one could always fall back to characters). They report that training with \emph{sampled} segmentation (termed ``subword regularization'') instead of just using one deterministic segmentation indeed improves machine translation performance.
The same motivation led \citet{provilkov-etal-2020-bpe} to propose BPE-dropout where the skipping of individual merges in the BPE construction leads to variety in segmentations.
Subword regularization not only has been shown to help in monolingual in-domain tasks, but also for improving transfer in multilingual models, see \cref{sec:multilingual}.

The observation that sampling segmentation helps is confirmed by \citet{hiraoka-etal-2019-stochastic}, who employ a Bayesian nonparametric model (see \cref{sec:nonparametrics}) as the LM that defines the tokenization.

\citet{DBLP:journals/corr/abs-2103-01421} build a similar model where the unigram LM is based on character-level BiLSTM encodings of the input and apply it to unsupervised Chinese Word Segmentation (see \cref{sec:chinesewordsegmentation}).%
\footnote{Note that this thus character-conditioned model can also be seen as an example of segmental neural language models (\cref{sec:segmental}).}

\subsubsection{SentencePiece \citep{kudo-richardson-2018-sentencepiece}}
Not itself an algorithm as often assumed, but actually a software package, SentencePiece \citep{kudo-richardson-2018-sentencepiece} offers both BPE and Unigram LM algorithms (so specifying ``SentencePiece'' is certainly not informative enough). Importantly, unlike their other implementations it does not treat spaces as special guaranteed word boundaries, allowing learned units to cross these boundaries and obviating the need for pre-tokenization in languages without whitespace-tokenized words like Chinese and Japanese.

\subsection{Comparing morphological segmentation to BPE and friends}
\label{comparing-morphological-segmentation-to-bpe}

Several studies have compared linguistically motivated segmentation with data-driven ones, without conclusive results (to say the least).

\citet{bostrom-durrett-2020-byte} claim that UnigramLM obtains better segmentation than BPE, both qualitatively (they tend to better correspond to morphemes and Morfessor \citep{10.1145/1187415.1187418} morphs) and quantitatively (they improve BERT-style models' performance on modern NLP tasks a little in English and a lot in Japanese).

When using (manually analyzed or gold) morphological analysis, \citet{MatNeuDye18Using} show that language modeling can be improved for agglutinative languages like Turkish.
In \citet{schwartz2020neural}'s low-resource study shows Morfessor-based language models (and character-based ones, see \cref{sec:purechar}) outperform BPE-based ones.
\citet{DBLP:journals/corr/abs-2001-01589} likewise improve NMT on Turkish and Uyghur by using morphological analyzers before applying BPE.

Using unsupervisedly obtained ``morphological'' subwords on the other hand, only \citet{ataman2018evaluation} find that a model based on Morfessor FlatCat can outperform BPE; \citet{zhou2018}, \citet{domingo2018}, \citet{machavcek2018morphological}, and \citet{saleva-lignos-2021-effectiveness} find no reliable improvement over BPE for translation.
\Citet{banerjee2018meaningless} analyze translations obtained segmenting with Morfessor and BPE, and conclude that a possible improvement depends on the similarity of the languages.
\citet{huck2017target} propose thus to combine both approaches.

As a possible explanation for the good performance of BPE, \citet{galle2019investigating} claims that the performance of BPE is related to its compression capacity: with respect to members from the same class of compression algorithm, BPE performs close to the top in data compression benchmarks.

\subsection{How many units do we need?}

Another open question is the question of how many merges (or what other prior) one should select for optimal performance. 
The best-performing number may depend on the task and the domain and differ by language \citep{mielke-etal-2019-kind,domingo2018much}.%
\footnote{Relatedly, \citet{DBLP:journals/corr/abs-2102-02585} show that the subword size matters and differs somewhat systematically between languages in the $n$-gram based fastText embeddings \citep{TACL999}.}
More and thus larger subword units allow for and lead to more memorization \citep{kharitonov2021bpe}, which may or may not be desired depending on the application.

Predicting the number of merges that works best without having to try different sizes would be desirable and \citet{gowda-may-2020-finding} claim to have found one such heuristic: merge as much as possible to shorten the overall sequence lengths while making sure that 95\% of subword units appear at least 100 times (presumably in training data). Their motivation is that neural machine translation models are frequency-biased in their outputs and thus maintaining a more uniform frequency distribution is better.%
\footnote{A somewhat similar approach is taken by \citet{gutierrez-vasques-etal-2021-characters}, who look at the entropy (and transformations) of the distribution over subword types to identify a ``turning point'' at which one of the transformed quantities is minimal---but this turning point happens with far fewer merges than are generally required to reach good performance.}
A similar study undertaken by \citet{ding-etal-2019-call} reiterate how contradictory suggestions for number of merges in past work are and add that in low-resource scenarios far fewer merges seem to be better, a trend with Transformers which differs from that with LSTMs, leading to an interesting question: should smaller corpora mean you can't afford characters or is it rather that you can't afford words?

A simple online answer to the question of how to select merges is presented by \citet{SalRunCod18Optimizing}: while training an NMT model using BPE segments, gradually \emph{increase} the vocabulary size by merging BPE vocabulary items, adding new, bigger BPE segments until they obtain diminishing returns. Embeddings for the newly introduced subwords are initialized by merging the embeddings of the two merged BPE segments with an autoencoder.
Formulating the vocabulary selection problem as a search for the set of tokens with the highest entropy, ~\citet{xu2021vocabulary} proposes an optimal transport driven selection from BPE units that obtains vocabulary merges that often outperform a language-independent standard setting for translation.
Another recent method that comes with a stopping criteria (and therefore dispenses with an additional hyperparameter) is \citet{vilar2021statistical} which defines the likelihood of a vocabulary with respect to a sequence, and improves that likelihood greedily.

\section{Shared vocabularies in multilingual models}\label{sec:multilingual}

Many NLP applications process text in more than one language at a time, the most obvious example perhaps being a machine translation system. In such cases, one could either use (and potentially learn) a tokenizer per language or a single tokenizer for both languages (also allowing sharing of embeddings if desired).
Building a highly multilingual system that translates between more than two languages, \citet{johnson-etal-2017-googles} perform the former and first encounter questions like whether to oversample low-resource languages for learning a data-driven tokenizer and if so to which degree.
These questions are addressed differently in the now more common highly multilingual pre-trained Transformers like mBERT \citep{DevChaLee18Bert} and XLM \citep{NEURIPS2019_c04c19c2}, and XLM-R \citep{conneau-etal-2020-unsupervised}. In these models the sharing of learned representations is hypothesized to help transfer between languages by \citet{pires-etal-2019-multilingual}, though \citet{wu-dredze-2019-beto} provide inclonclusive results for this claim.
It is worth pointing out that \citet{K2020Cross-Lingual} disagree and claim that subword overlap is not as important for transfer.

Even though all these models settle make sure to oversample low-resource languages at least some amount, \citet{acs2019exploring} and \citet{rust2021good} show that tokenization in BERT-based Transformers is still biased towards high-resource languages. This bias is visible in a word's ``fertility,'' i.e., the number of subwords a word is split into on average (which for example is much lower for English than it is for, say, Finnish), but they also find it affecting results in controlled (comparing monolingual to multilingual tokenization) downstream tasks.
\citet{maronikolakis-etal-2021-wine-v} find that these granularity differences in tokenization between languages also greatly affect sharing of semantic representations.

For selecting appropriate tokenization in a multilingual setting, \citet{chung-etal-2020-improving} offer an approach for retraining models from scratch, selecting subword vocabularies for language \emph{clusters} to explicitly control for allocation and sharing.
If on the other hand retraining from scratch is not feasible, one option is to add new subword units for the underresourced/oversegmented languages. \citet{wang-etal-2020-extending} and \citet{chau-etal-2020-parsing} both propose such additions with randomly initialized embeddings, but these approaches did not perform well when studied by \citet{ebrahimi2021adapt}; extending the idea, \citet{liu2021bridging} propose to use information about existing subword units to estimate embeddings instead of initializing newly added units randomly (similar to \citet{SalRunCod18Optimizing}).
A different option is proposed by \citet{wang-etal-2021-multi-view}, who instead force the model to use (already existing) smaller subword units in high-resource languages like English to make the segmentations across languages more similar and thus aid transfer---thus avoiding the complete retraining that comes with changing the segmentation method or allocation.
In particular, they fine-tune the model to perform the target task 1) well with the deterministic segmentation (that undersegments high-resource languages relative to low-resource ones), 2) well with sampled segmentations (so even high-resource languages' words are segmented more), and 3) equally (in terms of a low divergence between the respective output distributions) between the two.

\section{``Tokenization-free'' character- and byte-level modeling}\label{sec:purechar}

In sections \cref{sec:augmenting} and \cref{sec:related-twolevel} we discussed augmenting word models with character-level information in closed- and open-vocabulary settings. The idea of pure character- or byte-level modeling seems like an obvious simplification.
Indeed, \citet{SutMarHin11Generating} successfully modeled strings one character at a time using multiplicative RNNs, \citet{Chr13Text} suggest using character-level RNN modeling for agglutinative languages and tasks that require character information like character-level text segmentation (even anticipating contextualized embeddings!), and
\citet{conneau-etal-2017-deep} successfully perform text classification from raw characters.
The big breakthrough for generative character/byte-level models however came only with \citet{al2019character}, who showed that sufficiently deep Transformers (64 layers in their case) can \emph{greatly} outperform previous subword-based and hybrid (\cref{sec:related-twolevel}) open-vocabulary language models. This finding was updated by \citet{choe2019bridging}, who again manage to match previous word- and subword-based state-of-the-art language modeling results.

\subsection{Characters?}

A major factor limiting the adoption of character-level models is the fact that character sequences tend to be much longer than their word- or subword-level counterparts, making training and inference slower.
To improve training speed and efficiency, \citet{libovicky-fraser-2020-towards} propose to start with a subword-based model and then fine-tune that model to instead work over characters, though they find improvements only in one of two evaluated language pairs.
The more common approach to both training and inference however are various architectures for subsampling sequences, particularly in applications to machine translation:
\citet{chung2016character} introduce a bi-scale recurrent neural network to enable the decoder of an encoder-decoder model to produce character sequences; they demonstrate improved performance over a subword-level decoder.
\citet{TACL1051} (and later \citet{gao2020character}) advocate for the use of convolution and pooling layers at the input of the encoder.
\citet{CheFosBap18Revisiting} evaluate various temporal pooling strategies including the HM-RNN of \citet{ChuAhnBen17Hierarchical} (discussed in \cref{sec:segmental}) and conclude that none of them offered a suitable trade-off of performance and speed.

It has been argued that character-level models are more robust to noise and out-of-distribution data \citep{gupta2019character}, possibly because a word- or subword-level token sequence will change dramatically in the presence of noise.
\citet{libovicky2021dont} however survey multiple character-level MT systems and conclude that they ``show neither better domain robustness, nor better morphological generalization, despite being often so motivated,'' a finding corroborated by \citet{rosales-nunez-etal-2021-noisy} for noisy user-generated text.
Specifically under the lens of gender bias, \citet{gaido-etal-2021-split} argue that character-level processing can lead to less gender bias in models: data-driven BPE vocabularies are biased towards male forms, for one because of frequency, but also because in languages like French and Italian, female forms are often constructed by affixing male forms, leading to a disparity in tokenization. They show that character-level processing can ameliorate these disparities to some degree.

\subsection{Character hashes?}

In massively multilingual settings, naïve use of a character-level model can result in a very large vocabulary.
For example, Unicode 13.0 specifies 143,859 codepoints, most of which will never be seen in training.
This can lead to ``out-of-vocabulary'' problems just like the ones encountered with words (discussed in \cref{sec:word-level}).
One possible solution is used in \citet{clark2021canine}'s CANINE, where instead of giving each character its own embedding, they \emph{hash} all possible codepoints with multiple hash functions into smaller sets to share parameters across codepoints (similar to \citet{SveHanWin17Hash}).
CANINE further includes downsampling and upsampling operations to attain reasonable computational efficiency, and focuses on non-generative sequence labeling tasks.

\subsection{Bytes?}\label{sec:bytes}

An alternative way to avoid the huge-vocabulary problem of character-level models is to instead represent text using the byte sequence resulting from a standard encoding like UTF-8.
This can be seen as using a fixed tokenization scheme that was created by a standards body (the Unicode consortium) rather than a linguistically- or statistically-informed process.
The main advantage of using Unicode bytes specifically is their incredible scope -- the Unicode consortium states that their goal is to cover ``all the characters for all the writing systems of the world, modern and ancient'' in addition to ``technical symbols, punctuations, and many other characters used in writing text''.\footnote{\url{https://unicode.org/faq/basic_q.html}}
Separately, byte-level modeling is presented as a natural choice by \citet{graves2013generating} when following the ``principle of modelling the smallest meaningful units in the data''.
As shown recently by ByT5 \citep{xue2021byt5}, byte-level models confer similar benefits to character-level models (better robustness to noise, avoidance of out-of-vocabulary issues, etc.)\ with similar drawbacks (longer training and inference time due to longer sequences).
As such, similar results and models have been proposed for byte-level models as for character-level.
For example, the recent Charformer model \citep{tay2021charformer} downsamples input sequences to avoid increased computational costs, and \citet{al2019character} demonstrate strong language modeling performance with a deep byte-level Transformer.

\subsection{So are these maximal decompositions the solution then?}

While byte-level modeling is often presented as an unbiased, token-free approach, we argue it is more constructive to consider byte-level modeling as simply using a fixed, predefined, and standardized tokenization scheme (that incidentally is often the same as the way that underlying the text data is stored on disk).
This tokenization scheme is by no means the ``best'' or most fundamental -- indeed, the Unicode standard was not created with any linguistically-motivated goal in mind, apart from being able to represent a huge variety of content.
Indeed, using a Unicode-based byte-level tokenization scheme will result in significantly different trade-offs for different languages by virtue of the way the standard was created:
Latin (i.e.\ ASCII) characters are represented as a single byte, whereas characters in other languages are represented as multiple bytes.
An immediate consequence is that a UTF-8-based byte-level tokenization scheme can result in dramatically longer sequences when representing the same underlying semantic content in different languages.
This could unfairly increase computational costs for downstream users of a model who do not communicate in ASCII symbols.

\subsection{Visual featurization: Pixels?}

Another approach to ``tokenization-free'' modeling utilizes \emph{visual} rather than byte-based representations.
Where byte-level models aim to cover the full underlying `vocabulary' as represented \emph{on disk}, visual representations aim to encode similarities among characters which \emph{human readers} may use when processing text. 
One motivation is robustness: byte-level models are reliant on consistent observed sequences, which leaves them susceptible to variation and noise which may render similarly visually.\footnote{For languages without a digital orthographic standard (e.g., Pashto), multiple byte sequences render similarly and are in free variation among speakers.}

The initial motivation for much work on using visual features were to create embeddings that reflected the shared character components found in e.g., Chinese characters (radicals) or Korean syllable blocks, and so were better able to generalize to rare or unseen characters. 
The first work in this space initialized embeddings for Chinese by linearizing bitmaps of characters or words \cite{aldon-etal-2016-neural,costa-jussa-etal-2017-chinese}. 
Subsequent work focused on character segmentation, either through pre-computed embeddings based on linearized images \cite{wang2020word} or learned with the downstream task via convolutions \cite{liu-etal-2017-learning,dai2017glyph,broscheit-2018-learning,meng2019glyce}, improving results for rare characters. 
Character segmentation was in part motivated by application to Chinese and the focus on sub-character components, but also enabled the use of fixed-width images, a problem which \citet{sun-etal-2018-super,sun2019squared} instead tackled by rendering words into square images, wrapping characters as necessary. 

Recent work has proposed \emph{``tokenization-free''} visual representations \cite{mansimov-etal-2020-towards,salesky-etal-2021-robust}. 
Rather than rendering images for a given segmentation into words, characters, or bytes, and thus incorporating their previously discussed challenges, they render each sentence as an image and translate directly from pixel decompositions. \citet{salesky-etal-2021-robust} show that such models can perform competitively across a range of languages and scripts for machine translation and are significantly more robust to induced noise, including but not limited to unicode errors. 
\citet{mansimov-etal-2020-towards} explore pixel-to-pixel models; a challenging setting that is not yet competitive but neatly sidesteps many concerns with text-based models.

\section{Discussion and Conclusion}\label{sec:conclusion}

Tokenization, both the process and the term itself, has evolved significantly since the early days of NLP. In this paper, we traced their evolution and highlighted major changes over the years, connecting disparate intuitions.

Despite significant advancement, we have seen that there is no perfect method of handling tokenization. All options---from whitespace-delimited pre-tokens to learned subwords and down to bytes---have drawbacks and strengths.
Many desiderata may be fundamentally at odds with each other, e.g., the desire to decompose maximally for simple and robust processing with a desire to be computationally efficient in a way that is fair across languages---a question that particularly pertinent as the field turns its attention to greener NLP.
While there are applications for which characters and bytes or even pixels may be the tool of choice, there are a myriad of applications in which larger discrete tokens are desirable, both for interpretability and efficiency.
Recent work gives us new examples of this: \citet{zhang-etal-2021-ambert} improve BERT by feeding inputs tokenized multiple ways and \citet{dossou2021crowdsourced} use human-annotated larger-than-word vocabulary units to improve low-resource MT. \citet{itzhak2021models} show that using subword units need not mean giving up on spelling-information, showing that spellings can be recovered from subword-based pretrained models.

In conclusion, despite all the promises of neural networks of allowing ``end-to-endness'', tokenization is a clear testimony that that promised land is still far away.
Like often, this drawback is evident for practitioners who have to verify that the pre-trained models matches the tokenizer that is being used (a separated model in most NLP frameworks).
The fact that tokenization is handled completely independent of the down-stream tasks adds to this separation. 
As~\citet{henderson2020unstoppable} puts it:
``It remains to find effective neural architectures for learning the set of entities jointly with the rest of the neural model, and for generalising such methods from the level of character strings to higher levels of representation''.

\section*{Acknowledgements}

We would like to thank Shiyue Zhang, Vassilina Nikoulina, and Kamalkumar R for their notes, Kaustubh Dhole and Mike Tian-Jian Jiang for paper suggestions, and Jason Eisner for feedback on a draft. Samson is supported by Salesforce and Singapore's Economic Development Board under the Industrial Postgraduate Programme.

\bibliography{between}
\bibliographystyle{acl_natbib}

\end{document}